\documentclass{article}
\usepackage{kr}

\usepackage{times}
\usepackage{soul}
\usepackage{url}
\usepackage[hidelinks]{hyperref}
\usepackage[utf8]{inputenc}
\usepackage[small]{caption}
\usepackage{graphicx}
\usepackage{amsmath}
\usepackage{amsthm}
\usepackage{booktabs}
\usepackage{algorithm}
\usepackage{algorithmic}
\usepackage{tikz}
\usepackage[lofdepth,lotdepth]{subfig}

\usepackage{xcolor}
\newtheorem{example}{Example}
\newtheorem{theorem}{Theorem}
\newtheorem{definition}{Definition}
\newtheorem{proposition}[theorem]{Proposition}

\newcommand{\af}{{\cal F}}
\newcommand{\iaf}{{\cal I}}
\newcommand{\completion}{{\cal C}}
\newcommand{\args}{{\cal A}}
\newcommand{\uargs}{{\cal A}^?}
\newcommand{\atts}{{\cal R}}
\newcommand{\cf}{{\sf cf}}
\newcommand{\ad}{{\sf ad}}
\newcommand{\co}{{\sf co}}
\newcommand{\pr}{{\sf pr}}
\newcommand{\stb}{{\sf st}}
\newcommand{\gr}{{\sf gr}}

\title{Stability in Abstract Argumentation}

\author{%
Jean-Guy Mailly\and
Julien Rossit\\
\affiliations
LIPADE, Universit\'e de Paris\\
\emails
\{jean-guy.mailly, julien.rossit\}@u-paris.fr
}

\begin{document}

\maketitle

\begin{abstract}
The notion of stability in a structured argumentation setup characterizes situations where the acceptance status associated with a given literal will not be impacted by any future evolution of this setup. In this paper, we abstract away from the logical structure of arguments, and we transpose this notion of stability to the context of Dungean argumentation frameworks. In particular, we show how this problem can be translated into reasoning with Argument-Incomplete AFs. Then we provide preliminary complexity results for stability under four prominent semantics, in the case of both credulous and skeptical reasoning. Finally, we illustrate to what extent this notion can be useful with an application to argument-based negotiation.
\end{abstract}

\section{Introduction}
Formal argumentation is a family of non-monotonic reasoning approaches with applications to ({\em e.g.}) multi-agent systems \cite{argmas2011}, automated negotiation \cite{DimopoulosMM19} or decision making \cite{AmgoudV12}. Roughly speaking, we can group the research in this domain in two families: abstract argumentation \cite{Dung95} and structured argumentation \cite{BesnardGHMPST14}. The former is mainly based on the seminal paper proposed by Dung, where abstract argumentation frameworks (AFs) are defined as directed graphs where the nodes represent arguments and the edges represent attacks between them. In this setting, the nature of arguments and attacks is not defined, only their interactions are represented in order to determine the acceptability status of arguments. On the opposite, different settings have been proposed where the arguments are built from logical formulas or rules, and the nature of attacks is based on logical conflicts between the elements inside the arguments. See {\em e.g.} \cite{HOFA} for a recent overview of abstract and structured argumentation.

In a particular structured argumentation setting, the notion of stability has been defined recently \cite{TesterinkOB19}. Intuitively, it represents a situation where a certain argument of interest will not anymore have the possibility to change its acceptability status. Either it is currently accepted and it will remain so, or on the contrary it is currently rejected, and nothing could make it accepted in the future. In the existing work on this topic, the authors mention some application to crime investigation (more precisely, Internet trade fraud). We also have in mind some other natural applications, like automated negotiation. For instance, if an agent is certain her argument for supporting her preferred offer cannot be accepted at any future step of the debate, she can switch her offer to another one, that may be less preferred, but at least could be accepted.

In this paper, we adapt the notion of stability to abstract argumentation, and we show that checking stability is equivalent to performing some well-known reasoning tasks in Argument-Incomplete AFs \cite{BaumeisterRS15,BaumeisterNR18,NiskanenNJR20}. While existing work on stability in structured argumentation focuses on a particular semantics (namely the grounded semantics), our approach is generic with respect to the underlying extension-based semantics. Moreover we consider both credulous and skeptical variants of argumentative reasoning.

This paper is organized as follows.~Section~\ref{section:background} introduces the basic notions of abstract argumentation literature in which our work takes place, and presents the concept of stability for structured argumentation frameworks. We then propose in Section~\ref{section:stability} a counterpart of this notion of stability adapted to abstract argumentation frameworks, and we show how we can reduce it to well-known reasoning tasks. We provide some lower and upper bounds for the computational complexity of checking whether an AF is stable.
Section~\ref{section:nego} then describes an application scenario in the context of automated negotiation.
Finally, Section~\ref{section:related-work} discusses related work, and Section~\ref{section:conclusion} concludes the paper by highlighting some promising future works.

\section{Background}\label{section:background}
\subsection{Abstract Argumentation}
Let us first introduce the abstract argumentation framework defined in \cite{Dung95}.

\begin{definition}
An {\em argumentation framework} (AF) is a pair $\af = \langle \args, \atts\rangle$ where $\args$ is the set of {\em arguments} and $\atts \subseteq \args \times \args$ is the {\em attack relation}.
\end{definition}

In this framework, we are not concerned by the precise nature of arguments ({\em e.g.} their internal structure or their origin) and attacks ({\em e.g.} the presence of contradictions between elements on which arguments are built). Only the relations between arguments ({\em i.e.} the attacks) are taken into account to evaluate the acceptability of arguments.

We focus on finite AFs, {\em i.e.} AFs with a finite set of arguments. For $a,b \in \args$, we say that $a$ {\em attacks} $b$ if $(a,b) \in \atts$. Moreover, if $b$ attacks some $c \in \args$, then $a$ {\em defends} $c$ against $b$. These notions are extended to sets of arguments: $S \subseteq \args$ attacks (respectively defends) $b \in \args$ if there is some $a \in S$ that attacks (respectively defends) $b$. The acceptability of arguments is evaluated through a notion of extension, {\em i.e.} a set of arguments that are jointly acceptable. To be considered as an extension, a set has to satisfy some minimal requirements:
\begin{itemize}
\item $S \subseteq \args$ is conflict-free (denoted $S \in \cf(\af)$) iff $\forall a,b \in S$, $(a,b) \not \in \atts$;
\item $S \in \cf(\af)$ is admissible (denoted $S \in \ad(\af)$) iff $S$ defends all its elements against all their attackers.
\end{itemize}
  
Then, Dung defines several semantics:
\begin{definition}\label{def:semantics}
Given $\af = \langle \args, \atts \rangle$ an AF, a set $S \subseteq \args$ is:
\begin{itemize}
  \item a {\em complete} extension ($S \in \co(\af)$) iff $S \in \ad(\af)$ and $S$ contains all the arguments that it defends;
  \item a {\em preferred} extension ($S \in \pr(\af)$) iff $S$ is a $\subseteq$-maximal complete extension;
  \item the unique {\em grounded} extension ($S \in \gr(\af)$) iff $S$ is the $\subseteq$-minimal complete extension;
    \item a {\em stable} extension ($S \in \stb(\af)$) iff $S \in \cf(\af)$ and $S$ attacks each $a \in \args \setminus S$,
\end{itemize}
where $\subseteq$-maximal  and $\subseteq$-minimal  denote respectively the maximal and the minimal elements for classical set inclusion.
\end{definition}

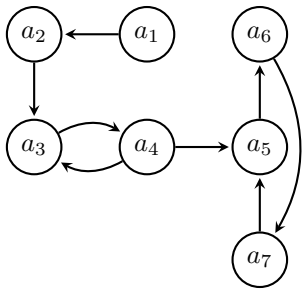
\begin{figure}[ht]
    \centering
    \begin{tikzpicture}[->,>=stealth,shorten >=1pt,auto,node distance=1.5cm, thick,main node/.style={circle,draw,font=\bfseries}]
\node[main node] (a1) {$a_1$};
\node[main node] (a2) [left of=a1] {$a_2$};
\node[main node] (a3) [below of=a2] {$a_3$};
\node[main node] (a4) [right of=a3] {$a_4$};
\node[main node] (a5) [right of=a4] {$a_5$};
\node[main node] (a6) [above of=a5] {$a_6$};
\node[main node] (a7) [below of=a5] {$a_7$};

\path[->] (a1) edge (a2) 
    (a2) edge (a3)
    (a3) edge[bend left] (a4)
    (a4) edge[bend left] (a3)
    (a4) edge (a5)
    (a5) edge (a6)
    (a6) edge[bend left] (a7)
    (a7) edge (a5);
\end{tikzpicture}
    \caption{An Example of AF $\af$}
    \label{fig:example-af}
\end{figure}

\begin{example}
Let $\af = \langle \args,\atts\rangle$ be the AF depicted in Figure~\ref{fig:example-af}. Nodes in the graph represent the arguments $\args$, while the edges correspond to the attacks $\atts$. Its extensions for $\sigma \in \{\gr,\stb,\pr,\co\}$ are given in Table~\ref{tab:example-af}.
\end{example}

\begin{table}[ht]
    \centering
    \begin{tabular}{l|c}
Semantics $\sigma$ & $\sigma$-extensions \\ \hline
grounded  & $\{\{a_1\}\}$\\
stable    & $\{\{a_1,a_4,a_6\}\}$ \\
preferred & $\{\{a_1,a_4,a_6\},\{a_1,a_3\}\}$ \\
complete  & $\{\{a_1,a_4,a_6\},\{a_1,a_3\},\{a_1\}\}$ \\
\end{tabular}
    \caption{$\sigma$-Extensions of $\af$}
    \label{tab:example-af}
\end{table}

We refer the interested reader to \cite{HOFASemantics} for more details about these semantics, as well as other ones defined after Dung's initial work. From the set of extensions $\sigma(\af)$ (for $\sigma \in \{\co,\pr,\gr,\stb\}$), we define two reasoning modes:
\begin{itemize}
\item an argument $a \in \args$ is credulously accepted with respect to $\sigma$ iff $a \in S$ for some $S \in \sigma(\af)$;
\item an argument $a \in \args$ is skeptically accepted with respect to $\sigma$ iff $a \in S$ for each $S \in \sigma(\af)$.
\end{itemize}

Then, a possible enrichment of Dung's framework consists in taking into account some uncertainty in the AF. This yields the notion of {\em Incomplete AFs}, studied {\em e.g.} in \cite{BaumeisterRS15,BaumeisterNR18,NiskanenNJR20}. Here, we focus on a particular type, namely {\em Argument-Incomplete AFs}, but for a matter of simplicity we just refer to them as Incomplete AFs.

\begin{definition}\label{def:iaf}
  An {\em incomplete argumentation framework} (IAF) is a tuple $\iaf = \langle \args, \uargs \atts\rangle$ where 
\begin{itemize}
\item $\args$ is the set of {\em certain arguments};
\item $\uargs$ is the set of {\em uncertain arguments};
\item $\atts \subseteq (\args \cup \uargs) \times (\args \cup \uargs)$ is the {\em attack relation};
\end{itemize}
and $\args, \uargs$ are disjoint sets of arguments.
\end{definition}

\begin{figure}[ht]
    \centering
    \begin{tikzpicture}[->,>=stealth,shorten >=1pt,auto,node distance=1.5cm, thick,main node/.style={circle,draw,font=\bfseries},uncertain/.style={circle,dotted,draw,font=\bfseries}]
\node[main node] (a1) {$a_1$};
\node[main node] (a2) [left of=a1] {$a_2$};
\node[main node] (a3) [below of=a2] {$a_3$};
\node[uncertain] (a4) [right of=a3] {$a_4$};
\node[main node] (a5) [right of=a4] {$a_5$};
\node[main node] (a6) [above of=a5] {$a_6$};
\node[uncertain] (a7) [below of=a5] {$a_7$};

\path[->] (a1) edge (a2) 
    (a2) edge (a3)
    (a3) edge[bend left] (a4)
    (a4) edge[bend left] (a3)
    (a4) edge (a5)
    (a5) edge (a6)
    (a6) edge[bend left] (a7)
    (a7) edge (a5);
\end{tikzpicture}
    \caption{An Example of IAF $\iaf$}
    \label{fig:example-iaf}
\end{figure}
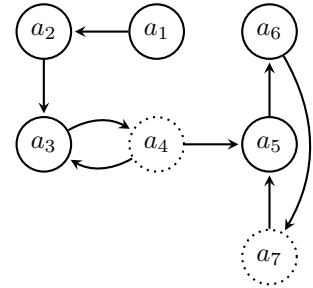

\begin{example}
The IAF $\iaf = \langle \args,\args^?,\atts\rangle$ is shown on Figure~\ref{fig:example-iaf}. The dotted nodes represent the uncertain arguments $\args^?$. Plain nodes and arrows have the same meaning as previously.
\end{example}

Uncertain arguments are those that may not actually belong to the system (for instance because of some uncertainty about the agent's environment). There are different ways to ``solve'' the uncertainty in an IAF, that correspond to different completions:
\begin{definition}
Given $\iaf = \langle \args, \uargs \atts\rangle$ an IAF, a {\em completion} is an AF $\af = \langle \args', \atts'\rangle$ where
\begin{itemize}
\item $\args \subseteq \args' \subseteq \args \cup \uargs$;
\item $\atts' = \atts \cap (\args' \times \args')$.
\end{itemize}
\end{definition}

\begin{figure}[ht]
    \centering
    \subfloat[$\completion_1$]{
    \centering
    \scalebox{0.8}{
    \begin{tikzpicture}[->,>=stealth,shorten >=1pt,auto,node distance=1.5cm, thick,main node/.style={circle,draw,font=\bfseries},uncertain/.style={circle,dotted,draw,font=\bfseries}]
\node[main node] (a1) {$a_1$};
\node[main node] (a2) [left of=a1] {$a_2$};
\node[main node] (a3) [below of=a2] {$a_3$};
\node (a4) [right of=a3] {};
\node[main node] (a5) [right of=a4] {$a_5$};
\node[main node] (a6) [above of=a5] {$a_6$};
\node (a7) [below of=a5] {};

\path[->] (a1) edge (a2) 
    (a2) edge (a3)
    (a5) edge (a6);
\end{tikzpicture}
} % End scalebox
} % End subfloat
\vline
\subfloat[$\completion_2$]{
\centering
    \scalebox{0.8}{
    \begin{tikzpicture}[->,>=stealth,shorten >=1pt,auto,node distance=1.5cm, thick,main node/.style={circle,draw,font=\bfseries},uncertain/.style={circle,dotted,draw,font=\bfseries}]
\node[main node] (a1) {$a_1$};
\node[main node] (a2) [left of=a1] {$a_2$};
\node[main node] (a3) [below of=a2] {$a_3$};
\node[main node] (a4) [right of=a3] {$a_4$};
\node[main node] (a5) [right of=a4] {$a_5$};
\node[main node] (a6) [above of=a5] {$a_6$};
\node (a7) [below of=a5] {};

\path[->] (a1) edge (a2) 
    (a2) edge (a3)
    (a3) edge[bend left] (a4)
    (a4) edge[bend left] (a3)
    (a4) edge (a5)
    (a5) edge (a6);
\end{tikzpicture}
} % End scalebox
} % End subfloat

\subfloat[$\completion_3$]{
    \centering
    \scalebox{0.8}{
    \begin{tikzpicture}[->,>=stealth,shorten >=1pt,auto,node distance=1.5cm, thick,main node/.style={circle,draw,font=\bfseries},uncertain/.style={circle,dotted,draw,font=\bfseries}]
\node[main node] (a1) {$a_1$};
\node[main node] (a2) [left of=a1] {$a_2$};
\node[main node] (a3) [below of=a2] {$a_3$};
\node (a4) [right of=a3] {};
\node[main node] (a5) [right of=a4] {$a_5$};
\node[main node] (a6) [above of=a5] {$a_6$};
\node[main node] (a7) [below of=a5] {$a_7$};

\path[->] (a1) edge (a2) 
    (a2) edge (a3)
    (a5) edge (a6)
    (a6) edge[bend left] (a7)
    (a7) edge (a5);
\end{tikzpicture}
} % End scalebox
} % End subfloat
\vline
\subfloat[$\completion_4$]{
\centering
    \scalebox{0.8}{
    \begin{tikzpicture}[->,>=stealth,shorten >=1pt,auto,node distance=1.5cm, thick,main node/.style={circle,draw,font=\bfseries},uncertain/.style={circle,dotted,draw,font=\bfseries}]
\node[main node] (a1) {$a_1$};
\node[main node] (a2) [left of=a1] {$a_2$};
\node[main node] (a3) [below of=a2] {$a_3$};
\node[main node] (a4) [right of=a3] {$a_4$};
\node[main node] (a5) [right of=a4] {$a_5$};
\node[main node] (a6) [above of=a5] {$a_6$};
\node[main node] (a7) [below of=a5] {$a_7$};

\path[->] (a1) edge (a2) 
    (a2) edge (a3)
    (a3) edge[bend left] (a4)
    (a4) edge[bend left] (a3)
    (a4) edge (a5)
    (a5) edge (a6)
    (a6) edge[bend left] (a7)
    (a7) edge (a5);
\end{tikzpicture}
} % End scalebox
} % End subfloat
    \caption{The Completions of $\iaf$}
    \label{fig:example-completions-iaf}
\end{figure}
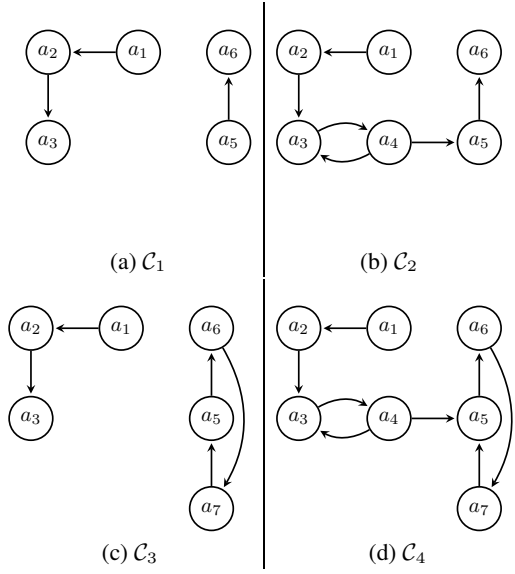

\begin{example}
Considering again $\iaf$ from the previous example, we show all its completions at Figure~\ref{fig:example-completions-iaf}. For each uncertain argument in $\args^? = \{a_4,a_7\}$, there are two possibilities: either the argument is present, or it is not. Thus, there are four completions.
\end{example}

This means that a completion is a ``classical'' AF made of all the certain arguments, some of the uncertain elements, and all the attacks that concern the selected arguments. Reasoning with an IAF generalizes reasoning with an AF, by taking into account either some or each completion. Formally, given $\iaf$ an IAF and $\sigma$ a semantics, the status of an argument $a \in \args$ is:
\begin{itemize}
\item {\em possibly credulously} accepted with respect to $\sigma$ iff $a$ belongs to some $\sigma$-extension of some completion of $\iaf$;
\item {\em possibly skeptically} accepted with respect to $\sigma$ iff $a$ belongs to each $\sigma$-extension of some completion of $\iaf$;
\item {\em necessarily credulously} accepted with respect to $\sigma$ iff $a$ belongs to some $\sigma$-extension of each completion of $\iaf$;
\item {\em necessarily skeptically} accepted with respect to $\sigma$ iff $a$ belongs to each $\sigma$-extension of each completion of $\iaf$.
\end{itemize}

\begin{example}
Let us consider again $\iaf$ from the previous example, and its completions $\completion_1$, $\completion_2$, $\completion_3$ and $\completion_4$. We observe that $a_1$ is necessarily skeptically accepted for any semantics, since it appears unattacked in every completion (thus, it belongs to every extension of every completion). 

On the opposite, $a_6$ is possibly credulously accepted with respect to the preferred semantics: it belongs to some extension of $\completion_4$. It is not skeptically accepted (because $\{a_1,a_3\}$ is a preferred extension of $\completion_4$ as well), and it is not necessarily accepted (because in $\completion_1$, it is not defended against $a_5$, thus it cannot belong to any extension).
\end{example}

\subsection{Stability in Structured Argumentation}
Now we briefly introduce the argumentation setting from \cite{TesterinkOB19}, based on \emph{ASPIC}\({}^{\mbox{+}}\) \cite{ModgilP14}.

Let us start with the notation that is used to represent the negation of a literal, {\em i.e.} for a propositional variable $p$, $-p = \neg p$ and $-(\neg p) = p$, with $\neg$ the classical negation. We call $p$ (respectively $\neg p)$ a positive (respectively negative) literal.

\begin{definition}
An argumentation setup is a tuple $AS = \langle L, R, Q,
K,\tau\rangle$ where:
\begin{itemize}
\item $L$ is a set of {\em literals} s.t. $l \in L$ implies
  $-l \in L$;
\item $R$ is a set of {\em defeasible rules} $p_1,\dots,p_m \Rightarrow q$ s.t. $p_1,\dots,p_m, q \in L$. Such a rule is called ``a rule for $q$''.
\item $Q \subseteq L$ is a set of {\em queryable literals}, s.t. no $q \in Q$ is a negative literal;
\item $K \subseteq L$ is the agent's (consistent) {\em knowledge base};
\item $\tau \in L$ is a particular literal called the {\em topic}.
\end{itemize}
\end{definition}

Usual mechanisms are used to define arguments and attacks. An argument for a literal $q$ is an inference tree rooted in a rule $p_1,\dots,p_m \Rightarrow q$, such that for each $p_i$, there is a child node that is either an argument for $p_i$, or an element of the knowledge base. Then, an argument $A$ attacks an argument $B$ if the literal supported by $A$ is the negation of some literal involved in the construction of $B$. From the sets of arguments and attacks built in this way, the grounded extension is defined as usual (see Definition~\ref{def:semantics}).

Given an argumentation setup $AS$, the status of the topic $\tau$ may be:
\begin{itemize}
\item {\em unsatisfiable} if there is no argument for $\tau$ in $AS$;
\item {\em defended} if there is an argument for $\tau$ in the grounded extension of $AS$;
\item {\em out} if there are some arguments for $\tau$ in $AS$, and all of them are attacked by the grounded extension;
\item {\em blocked} in the remaining case.
\end{itemize}

Then, stability can be defined, based on the following notion of future setups:

\begin{definition}
  Let $AS = \langle L, R, Q, K,\tau\rangle$ be an argumentation setup. The set of {\em future setups} of $AS$, denoted by $F(AS)$, is defined by $F(AS) = \{\langle L, R, Q, K',\tau\rangle \mid K \subseteq K'\}$. $AS$ is called {\em stable} if for each $AS' \in F(AS)$, the status of $\tau$ is the same as in $AS$.
\end{definition}

Intuitively, a future setup is built by adding new literals to the knowledge base (keeping the consistency property, of course). Then, new arguments and attacks may be built thanks to these new literals. The setup is stable if these new arguments and attacks do not change the status of the topic.

To conclude this section, let us mention that \cite{TesterinkOB19} provides a sound algorithm that approximates the reasoning task of checking the stability of the setup. This algorithm is however not complete, {\em i.e.} $AS$ is actually stable if the algorithm outcome is a positive answer, but there are stable setups that are not identified by the algorithm. This algorithm has the interest of being polynomially computable (more precisely, it stops in ${\cal O}(n^2)$ steps, where $n = |L| + |R|$).

\section{Stability in Abstract Argumentation}\label{section:stability}
In this section, we describe how we adapt the notion of stability to abstract argumentation. Contrary to previous works, we do not focus on a specific semantics, and thus we consider both credulous and skeptical reasoning. Moreover, we provide a translation of the stability problem into reasoning with AFs and IAFs. Despite being theoretically intractable, efficient algorithms exist for solving these problems in practice. So it paves the way to future implementations of an exact algorithm for checking stability, and its applications to concrete scenarios.

\subsection{Formal Definition of Stability in AFs}
From now on, we consider a finite {\em argumentation universe} that is represented by an AF $\af_U = \langle \args_U, \atts_U \rangle$. We suppose that any ``valid'' AF is made of arguments and attacks in $\af_U$, {\em i.e.} $\af = \langle \args, \atts\rangle$ s.t. $\args \subseteq \args_U$ and $\atts = \atts_U \cap (\args \times \args)$.

\begin{definition}
Given an AF $\af = \langle \args,\atts\rangle$, we call the {\em future AFs} the set of AFs $F(\af) = \{\af' = \langle \args', \atts'\rangle \mid \args \subseteq \args'\}$.
\end{definition}
Intuitively speaking, this means that a future AF represents a possible way to continue the argumentative process (by adding arguments and attacks), accordingly to $\af_U$. This corresponds to some kind of expansions of $\af$ \cite{BaumannB10}, where the authorized expansions are constrained by $\af_U$. This is reminiscent of the set of authorized updates defined in \cite{Saint-CyrBCL16}. Notice that $\af$ is a particular future AF.

Now we have all the elements to define stability.

\begin{definition}
Given an AF $\af = \langle \args,\atts\rangle$, $a \in \args$ an argument, and $\sigma$ a semantics, we say that $\af$ is credulously (respectively skeptically) $\sigma$-stable with respect to $a$ iff
\begin{itemize}
  \item either $\forall \af' \in F(\af)$, $a$ is credulously (respectively skeptically) accepted with respect to $\sigma$;
  \item or $\forall \af' \in F(\af)$, $a$ is not credulously (respectively skeptically) accepted with respect to $\sigma$.
\end{itemize}
\end{definition}

Although in this paper we focus on $\sigma \in \{\gr,\stb,\pr,\co\}$, the definition of stability is generic, and the concept can be applied when any extension semantics is used \cite{HOFASemantics}.

\begin{figure}[ht]
    \centering
    \subfloat[$\af_U$]{
    \centering
    \scalebox{0.8}{
    \begin{tikzpicture}[->,>=stealth,shorten >=1pt,auto,node distance=1.5cm, thick,main node/.style={circle,draw,font=\bfseries}]
\node[main node] (a1) {$a_1$};
\node[main node] (a2) [left of=a1] {$a_2$};
\node[main node] (a3) [below of=a2] {$a_3$};
\node[main node] (a4) [right of=a3] {$a_4$};
\node[main node] (a5) [right of=a4] {$a_5$};
\node[main node] (a6) [above of=a5] {$a_6$};
\node[main node] (a7) [below of=a5] {$a_7$};

\path[->] (a1) edge (a2) 
    (a2) edge (a3)
    (a3) edge[bend left] (a4)
    (a4) edge[bend left] (a3)
    (a4) edge (a5)
    (a5) edge (a6)
    (a7) edge (a5);
\end{tikzpicture}
} % End scalebox
} % End subfloat
\vline
\subfloat[$\af$]{
\centering
\scalebox{0.8}{
    \begin{tikzpicture}[->,>=stealth,shorten >=1pt,auto,node distance=1.5cm, thick,main node/.style={circle,draw,font=\bfseries}]
\node (a1) {};
\node (a2) [left of=a1] {};
\node[main node] (a3) [below of=a2] {$a_3$};
\node[main node] (a4) [right of=a3] {$a_4$};
\node[main node] (a5) [right of=a4] {$a_5$};
\node[main node] (a6) [above of=a5] {$a_6$};
\node[main node] (a7) [below of=a5] {$a_7$};

\path[->] (a3) edge[bend left] (a4)
    (a4) edge[bend left] (a3)
    (a4) edge (a5)
    (a5) edge (a6)
    (a7) edge (a5);
\end{tikzpicture}
} % End scalebox
} % End subfloat
    \caption{The Argumentation Universe $\af_U$ and a Possible AF $\af$}
    \label{fig:example-argumentation-universe}
\end{figure}
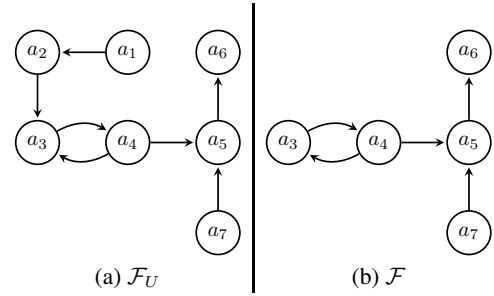

\begin{example}
Let us consider the argumentation universe $\af_U$ and the AF $\af$, both depicted in Figure~\ref{fig:example-argumentation-universe}. The argument $a_3$ is not credulously $\sigma$-stable for $\sigma = \stb$, since it is credulously accepted in $\af$, but not in the future AF where $a_2$ is added. On the contrary, it is skeptically $\sigma$-stable since it is not skeptically accepted in $\af$, nor in any future AF.

$a_6$ is skeptically $\sigma$-stable as well, but for another reason: indeed we observe that in $\af$ (and in any future AF), $a_6$ is defended by the (unattacked) argument $a_7$, thus it belongs to every extension.
\end{example}

On this simple example, it may seem obvious to determine that $a_5$, $a_6$ and $a_7$ will keep their status. However, let us notice that determining whether an argument keeps its status when an AF is updated has been studied, and is not a trivial question in the general case~\cite{BaroniGL14,AlfanoGP19}.

\subsection{Computational Issues}
We now provide a method for checking the stability of an AF with respect to some argument. The method is generic regarding the underlying extension semantics. It is based on the observation that the set of future AFs can be encoded into a single IAF (see Definition~\ref{def:iaf}).

\begin{definition}
  Given $\af = \langle \args, \atts \rangle$, the corresponding IAF is $\iaf_{\af} = \langle \args, \args_U \setminus \args, \atts_U \rangle$.
\end{definition}

The corresponding IAF is built from the whole set of arguments that appear in the universe. The ones that belong to $\af$ are the certain arguments, while the other ones are uncertain. Then of course, all the attacks from the universe appear in the IAF.  The set of completions of $\iaf_{\af}$ is actually $F(\af)$.

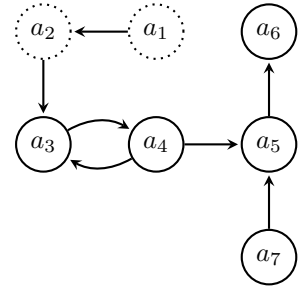
\begin{figure}[ht]
    \centering
     \begin{tikzpicture}[->,>=stealth,shorten >=1pt,auto,node distance=1.5cm, thick,main node/.style={circle,draw,font=\bfseries},uncertain/.style={circle,dotted,draw,font=\bfseries}]
\node[uncertain] (a1) {$a_1$};
\node[uncertain] (a2) [left of=a1] {$a_2$};
\node[main node] (a3) [below of=a2] {$a_3$};
\node[main node] (a4) [right of=a3] {$a_4$};
\node[main node] (a5) [right of=a4] {$a_5$};
\node[main node] (a6) [above of=a5] {$a_6$};
\node[main node] (a7) [below of=a5] {$a_7$};

\path[->] (a1) edge (a2) 
    (a2) edge (a3)
    (a3) edge[bend left] (a4)
    (a4) edge[bend left] (a3)
    (a4) edge (a5)
    (a5) edge (a6)
    (a7) edge (a5);
\end{tikzpicture}
    \caption{The IAF $\iaf_{\af}$ Corresponding to $\af$}
    \label{fig:corresponding-iaf}
\end{figure}

\begin{example}
Figure~\ref{fig:corresponding-iaf} shows the IAF corresponding to $\af$. The arguments that belong to the universe but not to $\af$ (namely, $a_1$ and $a_2$) appear as uncertain arguments. This means that the four completions of this IAF correspond to $F(\af)$.
\end{example}

We give a characterization of stability based on the IAF corresponding to an AF.

\begin{proposition}\label{prop:stability-af-iaf}
Given an AF $\af = \langle \args,\atts\rangle$, $a \in \args$ an argument, and $\sigma$ a semantics, $\af$ is credulously
(respectively skeptically) $\sigma$-stable with respect to $a$ iff
\begin{itemize}
  \item either $a$ is necessarily credulously (respectively skeptically) accepted in $\iaf_{\af}$ with respect to $\sigma$;
  \item or $a$ is not possibly credulously (respectively skeptically) accepted in $\iaf_{\af}$ with respect to $\sigma$.
\end{itemize}
\end{proposition}

This result shows that solving efficiently the stability problem is possible, using for instance the SAT-based piece of software \verb+taeydennae+ \cite{NiskanenNJR20} for reasoning in $\iaf_{\af}$.

Now, we provide preliminary complexity results. We start with upper bounds for the computational complexity of stability.

\begin{proposition}\label{prop:complexity-upper-bounds}
  The upper bound complexity of checking whether an AF is (credulously or skeptically) $\sigma$-stable with respect to an argument is as presented in Table~\ref{table:complexity-upper-bounds}.
\end{proposition}

\begin{table}[ht]
  \centering
  \begin{tabular}{c|c|c}
    $\sigma$ & Credulous & Skeptical \\ \hline
    $\stb$ & $\in \Pi_2^P$ & $\in \Sigma_2^P$ \\
    $\co$ & $\in \Pi_2^P$ & $\in$ coNP \\
    $\gr$ & $\in$ coNP & $\in$ coNP \\
    $\pr$ & $\in \Pi_3^P$ & $\in \Sigma_3^P$ \\
  \end{tabular}
  \caption{Upper Bound Complexity of Checking Stability \label{table:complexity-upper-bounds}}
\end{table}

\begin{proof}[Sketch of proof]
Non-deterministically guess a pair of future AFs $\af'$ and $\af''$. Check that $a$ is credulously (respectively skeptically) accepted in $\af'$, and $a$ is not credulously (respectively skeptically) accepted in $\af''$. The complexity of credulous (respectively skeptical) acceptance in AFs \cite{HOFAComplexity} allows to deduce an upper bound for credulous (respectively skeptical) stability.
\end{proof}

Now we also identify lower bounds for the computational complexity of stability.

\begin{proposition}\label{prop:complexity-lower-bounds}
  The lower bound complexity of checking whether an AF is (credulously or skeptically) $\sigma$-stable with respect to an argument is as presented in Table~\ref{table:complexity-lower-bounds}.
\end{proposition}

\begin{table}[ht]
  \centering
  \begin{tabular}{c|c|c}
    $\sigma$ & Credulous & Skeptical \\ \hline
    $\stb$ & NP-hard & coNP-hard \\
    $\co$ & NP-hard &  P-hard\\
    $\gr$ & P-hard & P-hard \\
    $\pr$ & NP-hard & $\Pi_2^P$-hard \\
  \end{tabular}
  \caption{Lower Bound Complexity of Checking Stability \label{table:complexity-lower-bounds}}
\end{table}

\begin{proof}[Sketch of proof]
Credulous (respectively skeptical) acceptance in an AF $\af$ can be reduced to credulous (respectively skeptical) stability, such that the current AF is $\af$, and the argumentation universe is $\af_U = \af$. Thus, $\af$ is credulously (respectively skeptically) $\sigma$-stable with respect to some argument $a$ iff $a$ is credulously (respectively skeptically) accepted in $\af$ with respect to $\sigma$. The nature of the reduction (its computation is bounded with logarithmic space and polynomial time) makes it suitable for determining both P-hardness and {\sf C}-hardness, for {\sf C} $\in \{$NP, coNP, $\Pi_2^P\}$. Thus, we can conclude that stability is at least as hard as acceptance in AFs. From known complexity results for AFs \cite{HOFAComplexity}, we deduce the lower bounds given in Table~\ref{table:complexity-lower-bounds}.
\end{proof}

\section{Applying Stability to Automated Negotiation}\label{section:nego}
Now, we discuss the benefit of stability in a concrete application scenario, namely automated negotiation. Let us consider a simple negotiation framework, where practical arguments ({\em i.e.} those that support some offers) are mutually exclusive, and for each agent there is a preference relation between the offers supported by these arguments (for instance, these preferences can be obtained from a notion of utility associated with each offer). So, each agent's goal is to make her preferred practical argument ({\em i.e.} the one that supports the preferred offer) accepted at the end of the debate. Each agent, in turn, can add one (or more) argument(s) that defend her preferred argument. In this first version of the negotiation framework, agents have a total ignorance about their opponent.

Then, an enriched version of this protocol can be defined, where the agents use the notion of argumentation universe to model their (uncertain) knowledge about the opponent. Then, stability can help the agent to obtain a better outcome: if at some point, the agent's preferred practical argument is rejected and stable, this means that this argument will not be accepted at the end of the debate, whatever the actual moves of the other agents. It is then profitable to the agent to change her goal, defending now the argument that supports her second preferred offer instead of the first one. This can reduce the number of rounds in the negotiation (and thus, any communication cost associated with these rounds), and even improve the outcome of the negotiation for the agent.

\begin{figure}[ht]
    \centering
    \begin{tikzpicture}[%->,>=stealth,
                        shorten >=1pt,auto,node distance=1.5cm,
                thick,main node/.style={circle,draw,font=\bfseries}]
\node[main node] (p1) {$p_1$};
\node[main node] (p2) [right of=p1] {$p_2$};
\node[main node] (p3) [right of=p2] {$p_3$};

\node[main node] (a1) [below left of=p1] {$a_1$};
\node[main node] (a2) [below of=p1] {$a_2$};
\node[main node] (a3) [below right of=p1] {$a_3$};

\path[->] (p1) edge (p2) 
    (p2) edge (p1)
    (p2) edge (p3)
    (p3) edge (p2)
    (p1) edge[bend left] (p3)
    (p3) edge[bend right] (p1)
    (a1) edge (p1)
    (a2) edge (p1)
    (a3) edge (p1);
\end{tikzpicture}
    \caption{The Negotiation Debate $\af_1$}
    \label{fig:negotiation-debate}
\end{figure}
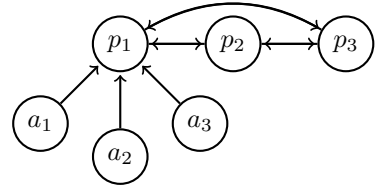

Let us now provide a concrete example. We suppose that the offers $O = \{o_1,o_2,o_3\}$ are supported by one practical argument each, {\em i.e.} $\{p_1,p_2,p_3\}$ with $p_i$ supporting $o_i$. The practical arguments are mutually exclusive. The preferences of the agents are opposed: agent $1$ has a preference ranking $o_3 >_1 o_2 >_1 o_1$, while the preferences of agent $2$ are \linebreak $o_1 >_2 o_2 >_2 o_3$. So, at the beginning of the debate, the goal of agent $1$ (respectively agent $2$) is to accept the argument $p_3$ (respectively $p_1$). Let us suppose that the first round consists in agent $1$ attacking the argument $p_1$ with three arguments $a_1, a_2$ and $a_3$, thus defending $p_3$. This situation is depicted in Figure~\ref{fig:negotiation-debate}.

In $\af_1$, that represents the state of the debate after agent $1$'s move, the argument $p_1$ is clearly rejected under the stable semantics,\footnote{As well as any semantics considered in this paper.} since it is not defended against $a_1, a_2$ and $a_3$. Consider that agent $2$ has one argument at her disposal, $a_4$,  with the corresponding attacks $(a_4,a_3)$ and $(a_4,p_2)$. Without a possibility to anticipate the evolution of the debate, the best action for agent $2$ is to utter this argument, thus defending $p_1$ against $a_3$ and $p_2$.

\begin{figure}[ht]
    \centering
    \begin{tikzpicture}[%->,>=stealth,
                        shorten >=1pt,auto,node distance=1.5cm,
                thick,main node/.style={circle,draw,font=\bfseries},uncertain/.style={circle,dotted,draw,font=\bfseries}]
\node[main node] (p1) {$p_1$};
\node[main node] (p2) [right of=p1] {$p_2$};
\node[main node] (p3) [right of=p2] {$p_3$};

\node[main node] (a1) [below left of=p1] {$a_1$};
\node[main node] (a2) [below of=p1] {$a_2$};
\node[main node] (a3) [below right of=p1] {$a_3$};

\node[uncertain] (a4) [below of=a3] {$a_4$};
\node[uncertain] (a5) [below of=a2] {$a_5$};
\node[uncertain] (a6) [below of=a1] {$a_6$};

\path[->] (p1) edge (p2) 
    (p2) edge (p1)
    (p2) edge (p3)
    (p3) edge (p2)
    (p1) edge[bend left] (p3)
    (p3) edge[bend right] (p1)
    (a1) edge (p1)
    (a2) edge (p1)
    (a3) edge (p1)
    (a4) edge (a3)
    (a4) edge[bend right] (p2)
    (a5) edge (a2)
    (a6) edge (a1)
    (a5) edge[bend left] (a6)
    (a6) edge[bend right] (a5);
\end{tikzpicture}
    \caption{The Argumentation Universe $\af_U$}
    \label{fig:argumentation-universe}
\end{figure}
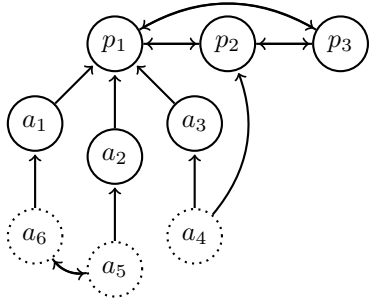

Now, let us suppose that agent $2$ has an opponent modelling in the form of the argumentation universe $\af_U$, described at Figure~\ref{fig:argumentation-universe}.

Now, we observe that $p_1$ does not appear in any extension of any future framework. Indeed, it is obvious that, if one of $a_4$, $a_5$ and $a_6$ is not present at the end of the debate, then $p_1$ is not defended against (respectively) $a_3$, $a_2$ or $a_1$. Otherwise, if $a_4$, $a_5$ and $a_6$ appear together, the mutual attack between $a_5$ and $a_6$ will be at the origin of two extensions, one where $a_6$ appears with $a_2$ (then defeating $p_1$), and the other one containing $a_5$ and $a_1$ (thus defeating again $p_1$). This means that $p_1$ is rejected in $\af_1$, and it is (both credulously and skeptically) $\sigma$-stable. In this situation, it is in the interest of agent $2$ to stop arguing, and proposing instead the option supported by the argument $p_2$. Indeed, according to the agent's preferences, $p_2$ is the best option if $p_1$ is not available anymore. Not only using the notion of stability in the argumentation universe allows to stop the debate earlier, but it also allows the agent $2$ to propose her second best option, which would not be possible if she had uttered $a_4$.

\section{Related Work}\label{section:related-work}
Dynamics of abstract argumentation frameworks \cite{DoutreM18} has received much attention in the last decade. We can summarize this field in two kinds of approaches: the goal either is to modify an AF to enforce some (set of) arguments as accepted, or to determine to what extent the acceptability of arguments is impacted by some changes in the AF. In the first family, we can mention extension enforcement \cite{BaumannB10}, that is somehow dual to stability. Enforcement is exactly the operation that consists in finding whether it is possible to modify an AF to ensure that a set of arguments becomes (included in) an extension, while stability is the property of an argument that will keep its acceptance status, whatever the future evolution of the AF. Control Argumentation Frameworks \cite{DimopoulosMM18} are also somehow related to stability, since they are a generalization of Dung's AFs that permit to realize extension enforcement under uncertainty.

The second family of works in the field of argumentation dynamics are those that propose efficient approaches to recompute the extensions or the set of (credulously/skeptically) accepted arguments when the AF is modified \cite{BaroniGL14,AlfanoGP19}. Although related to stability, these approaches do not provide an algorithmic solution to the problem studied in our work, since they focus on one update of the AF at once, instead of the set of all the future AFs.

\section{Conclusion}\label{section:conclusion}
In this paper, we have addressed a first study which investigates to what extent the notion of stability can be adapted to abstract argumentation frameworks.  
In particular, we have shown how it relates with Incomplete AFs, that are a model that integrates uncertainty in abstract argumentation. Our preliminary complexity results, as well as the translation of stability into reasoning with IAFs pave the way to the development of efficient computational approaches for stability, taking benefit from SAT-based techniques. Finally, we have shown that, besides the existing application of stability to Internet fraud inquiry \cite{TesterinkOB19}, this concept has other potential applications, like automated negotiation.

This paper opens the way for several promising research tracks. First of all, we plan to study more in depth complexity issues in order to determine tight results for the semantics that were studied here. Other direct future works include the investigation of other semantics, and the implementation of our stability solving technique in order to experimentally evaluate its impact in a context of automated negotiation.

We have focused on stability in extension semantics, which means that an argument will either remain accepted, or remain unaccepted. However, in some cases, it is important to deal more finely with unaccepted arguments. It is possible with $3$-valued labellings \cite{Caminada06}. Studying the notion of stability when such labellings are used to evaluate the acceptability of arguments is a natural extension of our work.

In some contexts, the assumption of a completely known argumentation universe is too strong. For such cases, it seems that using arbitrary IAFs (with also uncertainty on the attack relation) is a potential solution. Uncertainty on the existence (or direction) of attacks makes sense, for instance, when preferences are at play. Indeed, dealing with preferences in abstract argumentation usually involves a notion of defeat relation, that is a combination of the attacks and preferences. This defeat relation may somehow "cancel" or "reverse" the initial attack \cite{AmgoudC02,AmgoudV14,KaciTV18}, thus some uncertainty or ignorance about the other agents' preferences can be represented as uncertainty in the attack relation of the argumentation universe.

We are also interested in stability for other abstract argumentation frameworks. Besides preference-based argumentation that we have already mentioned, Dung's AFs has been generalized by adding a support relation \cite{AmgoudCLL08}, or associating quantitative weights with attacks \cite{DunneHMPW11} or arguments \cite{RossitMDM20}, or associating values with arguments \cite{Bench-Capon02}. But adapting the notion of stability to these frameworks may require different techniques than the one used in this paper. Also, the recent claim-based argumentation \cite{DvorakW20} provides an interesting bridge between structured argumentation and purely abstract frameworks. It makes sense to study stability in this setting, as a step that would make our results for different semantics and reasoning modes available for structured argumentation frameworks.

%% The file kr.bst is a bibliography style file for BibTeX 0.99c
\bibliographystyle{kr}
\bibliography{main}

\end{document}